\documentclass[10pt,journal]{IEEEtran}
\usepackage{amsmath,amssymb,epic,eepic,subfigure,multirow,epsfig}
\usepackage{mathrsfs,amsfonts}
\usepackage[dvips]{color}
\usepackage{algorithmic}
\usepackage{algorithm}
\usepackage{tabularx}
\usepackage{url}
\usepackage{graphicx}

\def\R{{\sl \hbox{I\kern-.2em\hbox{R}}}}

\def\si2{\sigma^2}

\def\sqr2{\frac{1}{\sqrt{2}}}

\begin{document}
\title{Online Adaptive Decision Fusion Framework Based on Entropic Projections onto Convex Sets with Application to Wildfire Detection in Video}

%
\author{\IEEEauthorblockN{Osman~G\"{u}nay\IEEEauthorrefmark{1},
Behcet~U\u{g}ur~T\"{o}reyin\IEEEauthorrefmark{2},
K{\i}van\c{c}~K\"{o}se\IEEEauthorrefmark{1} and
A.~Enis~\c{C}etin\IEEEauthorrefmark{1} \\
}
\IEEEauthorblockA{\IEEEauthorrefmark{1}Department of Electrical and Electronics Engineering\\
Bilkent University,
Ankara, Turkey, 06800\\ 
Telephone: 	+90-312-290-1219 Fax: 	+90-312-266-4192\\
Email: \{osman,kkivanc,cetin\}@ee.bilkent.edu.tr}\\
\IEEEauthorblockA{\IEEEauthorrefmark{2}Texas A\&M University at Qatar, Education City, 23874, Doha, Qatar 
\\
Telephone:+974-5508-9336\\
Email: behcet.toreyin@qatar.tamu.edu}
}

\maketitle
\textbf(EDICS:) ARS-IIU:Image \& Video Interpretation and Understanding
Object recognition and classification; Foreground/background segregation; Scene analysis

\begin{abstract}
In this paper, an Entropy functional based online Adaptive Decision Fusion (EADF) framework is developed for image analysis and computer vision applications. In this framework,
it is assumed that the compound algorithm consists of several sub-algorithms each of which yielding
its own decision as a real number centered around zero, representing the
confidence level of that particular sub-algorithm. Decision values
are linearly combined with weights which are updated online according to an
active fusion method based on performing entropic projections onto convex sets describing sub-algorithms. It is assumed that there is an oracle, who is usually a human operator, providing feedback to the decision fusion method. A video based wildfire detection system is developed to evaluate the performance of the algorithm in handling the problems where data arrives sequentially. In this case, the oracle is the security guard of the forest lookout tower verifying the decision of the combined algorithm. Simulation results are presented. The EADF framework is also tested with a standard dataset.
\end{abstract}


\section{Introduction}
\PARstart{A}{n} online learning framework called Entropy functional based Adaptive Decision Fusion (EADF) is proposed which can be used for various image analysis and computer vision applications. In this framework, it is assumed that the compound algorithm consists of several sub-algorithms each of which yielding its own decision. The final decision is taken based on a set of real numbers representing confidence levels of various sub-algorithms. Decision values are linearly combined with weights which are updated online using an active fusion method based on performing entropic projections (e-projections) onto convex sets describing sub-algorithms. 

Adaptive learning methods based on orthogonal projections are successfully used in some computer vision and pattern recognition problems~\cite{osman,Theodoridis}.
In this active learning approach decisions from different classifiers are combined using a linear combiner~\cite{dasarathy}. A multiple classifier system can prove useful for difficult pattern recognition problems especially when large class sets and noisy data are involved, because it allows the use of arbitrary feature descriptors and classification procedures at the same time~\cite{multipleclassifier}. Instead of determining the weights using orthogonal projections as in~\cite{osman,Theodoridis}, we introduce the entropic e-projection approach which is based on a generalized projection onto convex set.

The studies in the field of collective recognition, which were started in the middle of the 1950s, found wide application in practice during the last decade, leading to solution to complex large-scale applied problems~\cite{collectiverec}. One of the first examples of the use multiple classifiers was given by Dasarathy in~\cite{dasarathy} in which he introduced the concept  of  composite classifier systems as  a means of  achieving  improved  recognition system performance compared to employing the  classifier  components individually.  The method is illustrated by  studying  the  case  of  the  linear/NN(Nearest Neighbor)  classifier  composite system. Kumar and Zhang used multiple classifiers for palmprint recognition by characterizing the user's identity through the simultaneous use of three major palmprint representations and achieve better performance than either one individually~\cite{Kumar2005}. A multiple classifier fusion algorithm is proposed for developing an effective video-based face recognition method~\cite{facemultiple}. Garcia and Puig present results showing that pixel-based texture classification can be significantly improved by integrating texture methods from multiple families, each evaluated over multisized windows~\cite{texturemultiple}. This technique consists of an initial training stage that evaluates the behavior of each considered texture method when applied to the given texture patterns of interest over various evaluation windows of different size.


In this article, the EADF framework is applied to a computer vision based wildfire detection problem. The system based on this method is currently being used in more than 50 forest fire lookout towers. The proposed automatic video based wildfire detection algorithm is based on five sub-algorithms: (i)~slow moving video object detection, (ii)~smoke-colored region detection, (iii)~wavelet transform based region smoothness detection, (iv)~shadow detection and elimination, (v)~covariance matrix based classification. Each sub-algorithm decides on the existence of smoke in the viewing range of the camera separately. Decisions from sub-algorithms are combined
together by the adaptive decision fusion method. Initial weights of the sub-algorithms are determined from actual forest fire videos and test fires. They are updated by using entropic e-projections onto hyperplanes defined by the fusion weights. It is assumed that there is an oracle monitoring the decisions of the combined algorithm. In the wildfire detection case, the oracle is the security guard. Whenever a fire is detected the decision should be acknowledged by the security guard. The decision algorithm will also produce false alarms in practice. Whenever an alarm occurs the system asks the security guard to verify its decision. If it is incorrect the weights are updated according to the decision of the security guard. The goal of the system is not to replace the security guard but to provide a supporting tool to help him or her. The attention span of a typical security guard is only 20 minutes in monitoring stations. It is also possible to use feedback at specified intervals and run the algorithm autonomously at other times. For example, the weights can be updated when there is no fire in the viewing range of the camera and then the system can be run without feedback.

The paper is organized as follows:
Entropy functional based Adaptive Decision Fusion (EADF) framework is described in
Section~\ref{sec:weight_adaptation}. The first part of the section describes our previous weight update algorithm which is obtained by orthogonal projections onto convex sets~\cite{osman}, the second part proposes entropy based e-projection method for weight update of the sub-algorithms. Section~\ref{sec:wildfire_detection} introduces the video based wildfire detection problem. The proposed framework is not restricted to the wildfire detection problem. It can be also used in other real-time intelligent video analysis applications in which a security guard is available. In Section~\ref{sec:building_blocks} each one of the five sub-algorithms which make up the compound (main) wildfire detection algorithm is described.  In Section~\ref{sec:experimental}, experimental results are presented and the proposed online active fusion method is compared with the universal linear predictor and the weighted majority algorithms. The proposed EADF method is also evaluated on a dataset from the UCI machine learning repository~\cite{uci}. Well-known classifiers (SVM, K-NN) are combined using EADF. During the training stage individual decisions of classifiers are used to find the weight of each classifier in the composite EADF classifier. Finally, conclusions are drawn in Section~\ref{sec:conclusion}.

\section{Adaptive Decision Fusion (ADF) Framework}
\label{sec:weight_adaptation} 

Let the compound algorithm be composed of $M$-many detection
sub-algorithms: $D_1,...,D_M$. Upon receiving a sample input $x$ at time step $n$, each
sub-algorithm yields a decision value $D_i(x,n)\in\mathbb{R}$ centered around zero. If $D_i(x,n) >0 $, it means that the event is detected by the $i$-th sub-algorithm. Otherwise, it is assumed that the event did not happen.
The type of the sample input $x$ may vary depending on the
algorithm. It may be an individual pixel, or an image region, or the
entire image depending on the sub-algorithm of the computer vision
problem. For example, in the wildfire detection problem presented in Section~\ref{sec:wildfire_detection}, the number of sub-algorithms is $M$=5 and each pixel at the location $x$ of incoming image frame is considered as a sample input for every detection algorithm.

Let ${\bf{{D}}}(x,n)=[D_1(x,n),...,D_M(x,n)]^T$, be the vector of decision values of the sub-algorithms for the pixel at location $x$ of input image frame at time step $n$, and ${\bf{w}}(x,n)=[w_1(x,n),...,w_M(x,n)]^T$ be the current weight vector. For simplicity we will drop $x$ in ${\bf{w}}(x,n)$ for the rest of the paper. 

We define
\begin{equation}\label{eq:y_hat_lms}
\hat{y}(x,n) = {\bf{{D^T}}}(x,n){\bf{w}}(n) = \sum_{i}
w_i(n)D_i(x,n)
\end{equation}
as an estimate of the correct classification result $y(x,n)$ of the
oracle for the pixel at location $x$ of input image frame at time
step $n$, and the error $e(x,n)$ as $e(x,n) = y(x,n) - \hat{y}(x,n)$. As it can be seen in the next subsection, the main advantage of the proposed algorithm compared to other related
methods in~\cite{WMA}\nocite{tsmc_xu92}\nocite{tsmc_kuncheva02}-\cite{tsmc_ParikhP07}, is the
controlled feedback mechanism based on the error term. Weights of
the algorithms producing incorrect (correct) decision is reduced
(increased) iteratively at each time step. Another advantage of the proposed algorithm is that it does not assume any specific probability distribution about the data.

\subsection{Set Theoretic Weight Update Algorithm based on Orthogonal Projections}
In this subsection, we first review the orthogonal projection based weight update scheme~\cite{osman}.
Ideally, weighted decision values of sub-algorithms should be equal
to the decision value of $y(x,n)$ the oracle:
\begin{equation}\label{eq:hyperplane_1}
y(x,n) = {\bf{{D}}}^T(x,n){\bf{w}}
\end{equation}
which represents a hyperplane in the M-dimensional space,
$\mathbb{R}^M$. Hyperplanes are closed and convex in $\mathbb{R}^M$. At time instant $n$, ${\bf{{D}}}^T(x,n){\bf{w}}(n)$
may not be equal to $y(x,n)$. In our approach, the next set of weights are determined
by projecting the current weight vector ${\bf{w}}(n)$ onto the
hyperplane represented by Eq.~\ref{eq:hyperplane_1}. The orthogonal
projection ${\bf{w}}(n+1)$ of the vector of weights ${\bf{w}}(n)\in
\mathbb{R}^M$ onto the hyperplane $y(x,n)={\bf{{D}}}^T(x,n){\bf{w}}$
is the closest vector on the hyperplane to the vector ${\bf{w}}(n)$.

Let us formulate the problem as a minimization problem:
\begin{equation}
\begin{array}{l}
\label{eq:app_ortho_proj}
~~~~~~\displaystyle{\mathop{\min}_{\bf{w^*}}} ||{\bf{w^*}} - {\bf{w}}(n)|| \\
\mbox{subject to } {\bf{{D}}}^T(x,n){\bf{w^*}}=y(x,n)
 \end{array}
\end{equation}
The solution can be obtained by using Lagrange multipliers. If we define
the next set of weights as $\bf{w}(n+1)=\bf{w^*}$ it can be obtained by the following iteration:
\begin{equation}\label{eq:app_weights_w_lambda}
{\bf{w}}(n+1) = {\bf{w}}(n) + \frac{\lambda}{2}{\bf{{D}}}(x,n)
\end{equation}
where the Lagrange multiplier, $\lambda$, can be obtained from the
hyperplane equation:
\begin{equation}
{\bf{{D}}}^T(x,n){\bf{w^*}} - y(x,n) = 0
\end{equation}
as follows:
\begin{equation}\label{eq:app_lambda}
\lambda = 2\frac{y(x,n) - \hat{y}(x,n)}{||{\bf{{D}}}(x,n)||^2} =
2\frac{e(x,n)}{||{\bf{{D}}}(x,n)||^2}
\end{equation}
where the error, $e(x,n)$, is defined as $e(x,n) = y(x,n) -
\hat{y}(x,n)$ and $\hat{y}(x,n) = {\bf{{D}}}^T(x,n){\bf{w}}(n)$.
Plugging this into~Eq.~\ref{eq:app_weights_w_lambda}
\begin{equation}\label{eq:app_weights_}
{\bf{w}}(n+1) = {\bf{w}}(n) +
\frac{e(x,n)}{||{\bf{{D}}}(x,n)||^2}{\bf{{D}}}(x,n)
\end{equation}
is obtained. Hence the projection vector is calculated according to Eq.~\ref{eq:app_weights_}.

Whenever a new input arrives, another hyperplane based on the new
decision values ${\bf{{D}}}(x,n)$ of sub-algorithms, is defined in
$\mathbb{R}^M$
\begin{equation}\label{eq:hyperplane_2}
y(x,n+1) = {\bf{{D}}}^T(x,n+1){\bf{w^*}}
\end{equation}
This hyperplane will not be the same as $y(x,n) =
{\bf{{D}}}^T(x,n){\bf{w}}(n)$ hyperplane. The next set of weights, ${\bf{w}}(n+2)$,
are determined by projecting ${\bf{w}}(n+1)$ onto the hyperplane in
Eq.~\ref{eq:hyperplane_2}. Iterated weights converge to the
intersection of hyperplanes~\cite{cetin_pocs},~\cite{Combettes}. The rate of convergence can be adjusted by introducing a relaxation parameter $\mu$ to Eq.~\ref{eq:app_weights_} as follows
\begin{equation}\label{eq:app_weightsmu}
{\bf{w}}(n+1) = {\bf{w}}(n) +
\mu\frac{e(x,n)}{||{\bf{{D}}}(x,n)||^2}{\bf{{D}}}(x,n)
\end{equation}
where $0<\mu<2$ should be satisfied to guarantee the convergence according to the projections onto convex sets~(POCS)
theory~\cite{pocs_theory}\nocite{cetin_pocs2}\nocite{pocs_theory}\nocite{Theodoridis1}-\cite{Wornell}.

If the intersection of hyperplanes is an empty set, then the updated
weight vector simply satisfies the last hyperplane equation. In
other words, it tracks decisions of the oracle by assigning proper
weights to the individual sub-algorithms~\cite{cetin_pocs2},~\cite{Theodoridis1}.

The relation between support vector machines and orthogonal projections onto halfplanes was established in~\cite{Theodoridis1},~\cite{Theodoridis2} and~\cite{Theodoridis3}. As pointed out in~\cite{Theodoridis2} SVM is very successful in batch settings but it cannot handle online problems with drifting concepts in which the data arrive sequentially.

\begin{algorithm}[H]
\begin{algorithmic}
\STATE{Adaptive Decision Fusion(x,n)} 
\FOR{$i$ = 1 to M}
\STATE{$w_i(0)$ = $\frac{1}{M},  Initialization$} 
\ENDFOR
\STATE{$e(x,n) = y(x,n) - \hat{y}(x,n) $} 
\FOR{$i$ = 1 to M}
\STATE{$w_i(n+1)\leftarrow w_i(n) + \mu
\frac{e(x,n)}{||{\bf{{D}}}(x,n)||^2}D_i(x,n) $}
\ENDFOR
\STATE{$\hat{y}(x,n) = \sum_{i} w_i(n)D_i(x,n)$} 
\IF{$\hat{y}(x,n)
\geq 0$}
\STATE{return 1} 
\ELSE 
\STATE{return -1}
\ENDIF
\end{algorithmic}
\caption{The pseudo-code for the POCS based algorithm}
\label{fig:algo_1}
\end{algorithm}

\subsection{Entropic Projection (E-Projection) Based Weight Update Algorithm}
\label{Algorithm}
The $l_1$ norm based minimization approaches provide successful signal reconstruction results in compressive sensing problems~\cite{baraniuk,candes, osher, osher2}. However the $l_0$ and $l_1$ norm based cost functions used in compressive sensing problems are not differentiable everywhere. The entropy functional approximates the $l_1$ norm $\sum_i{|w_i(n)|}$ for $w_i(n)>0$~\cite{Bregman}. Therefore it can be used to find approximate solutions to the inverse problems defined in
\cite{baraniuk,candes} and other application requiring $l_1$ norm minimization. Bregman developed convex optimization algorithms in 1960's and his algorithms are widely used in many signal reconstruction and inverse problems \cite{pocs_theory, Herman, Lent, Trussell, Sezan, Combettes, enis2,enis2,enis4, Theodoridis}. Bregman's method provides  globally convergent iterative algorithms for problems with convex, continuous and differentiable cost functionals $g(.)$:
\begin{equation}
\min_{\mathbf{w}\in C} \; {g(\mathbf{w})}
\label{Dg}
\end{equation}
such that
\begin{equation}
 \mathbf{D}^T(x,n) \mathbf{w}(n) =y   \mbox{ for each time index $n$}
 \label{hyperplane}
\end{equation}
In the EADF framework the cost function is $g(\mathbf{w})=\sum_i^M{w_i(n)log(w_i(n))}$ and each equation in (\ref{hyperplane}) represents a hyperplane $H_n \in \mathbb{R}^M$ which are closed and convex sets. In Bregman's method the iterative algorithm starts with an arbitrary initial estimate and successive e-projections are performed onto the hyperplanes $H_n,$ $n=1,2,...,N$  in each step of the iterative algorithm. 

The e-projection onto a closed and convex set is a generalized version of the orthogonal projection onto a convex set~\cite{Bregman}. Let  
 $\mathbf{w}(n)$ denote the weight vector for the $n_{th}$ sample. Its e-Projection $\mathbf{w}^*$ onto a closed convex set $C$ with respect to a cost functional $g(\mathbf{w})$ is defined as follows  
\begin{equation}
\mathbf{w}^* = \arg \min_{\mathbf{w}\in C} \; L(\mathbf{w},\mathbf{w}(n))
\label{Dproj}
\end{equation}
where
\begin{equation}
L(\mathbf{w},\mathbf{w}(n)) = g(\mathbf{w})-g(\mathbf{w}(n))-<\bigtriangledown g(\mathbf{w}), \mathbf{w}-\mathbf{w}(n)>
\label{Dproj1}
\end{equation}
In the adaptive learning problem, we have the hyperplane $H:
 \mathbf{D}^T(x,n).\mathbf{w}(n+1) =y(x,n) \quad$ for each sample $x$.
For each hyperplane $H$, the e-projection (\ref{Dproj}) is equivalent to 
\begin{eqnarray}
\bigtriangledown g(\mathbf{w}(n+1)) = \bigtriangledown g(\mathbf{w}(n)) + \lambda \mathbf{D(x,n)} \\
\mathbf{D}^T(x,n).\mathbf{w}(n+1) =y(x,n) 
\label{eq15}
\end{eqnarray}
where $\lambda$ is the Lagrange multiplier. As pointed above
the e-projection is a generalization of the orthogonal projection. When the cost functional is the Euclidean cost functional $g(\mathbf{w})= \sum_i {w}_i(n)^2$ the distance $L(\mathbf{w_1}, \mathbf{w_2})$ becomes the $l_2$ norm square of the difference vector $(\mathbf{w_1}-\mathbf{w_2})$, and the e-projection simply becomes the well-known orthogonal projection onto a hyperplane.

\begin{figure}[htb]
\begin{center}
\fbox{\includegraphics*[width=8.0cm]{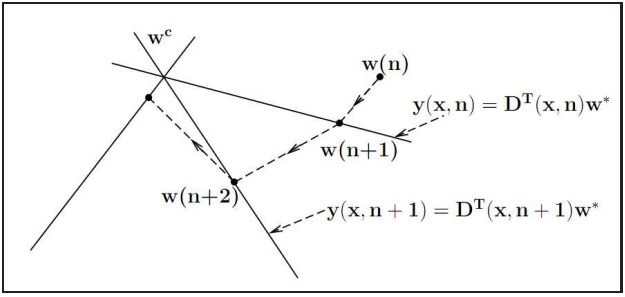}}
 \caption{Geometric interpretation of the entropic projection method: Weight vectors corresponding to
 decision functions at each frame are updated as to satisfy the hyperplane equations defined by the oracle's decision $y(x,n)$ and the decision vector
 ${\bf{D}}(x,n)$. Lines in the figure represent hyperplanes in $\mathbb{R}^M$. Weight update vectors converge to the intersection of the hyperplanes. Notice that e-projections are not orthogonal projections.}\label{fig:geo_int}
\end{center}
\end{figure}

When the cost functional is the entropy functional $g(\mathbf{w})= \sum_i {w}_i(n) \log({w}_i(n))$, the e-projection onto the hyperplane $H$ leads to the following update equations:
\begin{equation}
{w}_i(n+1) = {w}_i(n)e^{\lambda {D_i}(x,n)},\; i = 1,2,...,M
\label{entropicupdate1}
\end{equation}
where the Lagrange multiplier $\lambda$ is obtained by inserting (\ref{entropicupdate1}) into the hyperplane equation:
\begin{equation}
 \mathbf{D}^T(x,n)\mathbf{w}(n+1) =y(x,n)
\label{entropicupdate2}
\end{equation}
because the e-projection $\mathbf{w}(n+1)$ must be on the hyperplane $H$ in Eq.~\ref{eq15}. This globally  convergent iterative process is depicted in Fig.~\ref{fig:geo_int}. 

The above set of equations are used in signal reconstruction from Fourier Transform samples and the tomographic reconstruction problem~\cite{Herman,cetin_pocs2}. The entropy functional is defined only for positive real numbers which coincides with our positive weight assumption.

The pseudo-code for the e-projection based adaptive decision fusion based algorithm is given in Algorithm~\ref{fig:algent}. To find the $\lambda$ value that minimizes the squared error at each iteration a simple search between possible $\lambda$ values can be performed or a nonlinear equation should be solved (Eqs.~\ref{entropicupdate1} and~\ref{entropicupdate2}). 

\begin{algorithm}[!ht]
\begin{algorithmic}
\STATE{E-Projection Based Adaptive Decision Fusion(x,n)} 
\FOR{$i$ = 1 to M}
\STATE{$w_i(0)$ = $\frac{1}{M},  Initialization$} 
\ENDFOR

\FOR{$\lambda$ = $\lambda_{min}$ to $\lambda_{max}$}
\FOR{$i$ = 1 to M}
\STATE{$v_i(n) = w_i(n)$}
\STATE{$v_i(n+1)\leftarrow v_i(n) e^{\lambda D_i(x,n)}$}
\ENDFOR
\IF{ $||y(x,n)-\sum_{i} v_i(n+1)D_i(x,n)|| < ||y(x,n)-\sum_{i} v_i(n)D_i(x,n)||$ }
\STATE{$\mathbf{w_T}(n+1) \leftarrow \mathbf{v}(n+1)$}
\ENDIF
\ENDFOR
\STATE{$\mathbf{w}(n+1) \leftarrow \mathbf{w_T}(n+1)$}
\STATE{$\hat{y}(x,n) = \sum_{i} w_i(n)D_i(x,n)$} 
\IF{$\hat{y}(x,n)
\geq 0$}
\STATE{return 1} 
\ELSE 
\STATE{return -1}
\ENDIF
\end{algorithmic}
\caption{The pseudo-code for the EADF algorithm}
\label{fig:algent}
\end{algorithm}

\section{An Application: Computer Vision Based Wildfire Detection}
\label{sec:wildfire_detection} 
The Entropy function based Adaptive Decision Fusion (EADF) framework described in detail in the previous section with tracking capability is especially useful when the online active learning problem is of dynamic nature
with drifting concepts~\cite{schlimmer}\nocite{Polikar_1}-\cite{SMC_Conf3}. In video based wildfire detection problem introduced in this section, the nature
of forestal recordings vary over time due to weather conditions and
changes in illumination which makes it necessary to deploy an
adaptive wildfire detection system. It is not feasible to develop
one strong fusion model with fixed weights in this setting with
drifting nature. An ideal online active learning mechanism should
keep track of drifts in video and adapt itself accordingly. The
projections in Eq.~\ref{entropicupdate1} and 
Eq.~\ref{eq:app_weights_} adjust the importance of individual sub-algorithms by updating the
weights according to the decisions of the oracle.

Manned lookout posts are widely available in forests all around the
world to detect wild fires. Surveillance cameras can be placed in these surveillance towers to monitor the surrounding forestal area
for possible wild fires. Furthermore, they can be used to monitor
the progress of the fire from remote centers.

As an application of EADF, a computer vision based method for wildfire
detection is presented in this article. Security guards have to work
24 hours in remote locations under difficult circumstances. They may simply
get tired or leave the lookout tower for various reasons. Therefore,
computer vision based video analysis systems capable of producing
automatic fire alarms are necessary to help the security guards to reduce the average forest
fire detection time.

Cameras, once installed, operate at
forest watch towers throughout the fire season for about six months which is 
mostly dry and sunny in Mediterranean region. There is usually a guard in charge of the cameras, as well. The
guard can supply feed-back to the detection algorithm after the
installation of the system. Whenever an alarm is issued, she/he can
verify it or reject it. In this way, she/he can participate to the
learning process of the adaptive algorithm. The proposed active
fusion algorithm can be also used in other supervised learning
problems where classifiers combinations through feedback is required.

As described in the following section, the main wildfire detection
algorithm is composed of five sub-algorithms. Each algorithm has its
own decision function yielding a zero-mean real number for slow
moving regions at every image frame of a video sequence. Decision
values from sub-algorithms are linearly combined and weights of
sub-algorithms are adaptively updated in our approach.

There are several approaches on automatic forest fire detection 
in the literature. Some of the approaches are directed towards
detection of the flames using infra-red and/or visible-range cameras
and some others aim at detecting the smoke due to
wildfire~\cite{ollero2008}\nocite{Li2005}\nocite{bosch2007}\nocite{demirel}-\cite{fransizlarOE}. There are recent papers on sensor based fire detection~\cite{Hefeeda}\nocite{Sahin}-\cite{SMC_Conf1}. Infrared
cameras and sensor based systems have the ability to capture the
rise in temperature however they are much more expensive compared to
regular pan-tilt-zoom (PTZ) cameras. An intelligent space framework is described for indoor fire detection in~\cite{SMC_Conf2}. However, in this paper, an outdoor (forest) wildfire detection method is proposed.

\begin{figure}[ht]
\centerline{\includegraphics*[width=8.5cm]{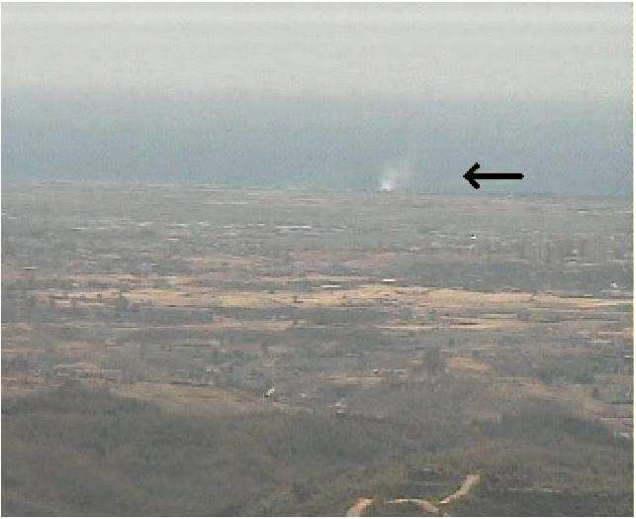}}
  \caption{Snapshot of a typical wildfire smoke captured by a forest watch tower which is 5~km away from the fire
(rising smoke is marked with an arrow).} \label{fig:snapshot}
\end{figure}

It is almost impossible to view flames of a wildfire from a camera
mounted on a forest watch tower unless the fire is very near to the
tower. However, smoke rising up in the forest due to a fire is
usually visible from long distances. A snapshot of a typical
wildfire smoke captured by a lookout tower camera from a distance
of 5~km is shown in Fig.~\ref{fig:snapshot}.

Guillemant and Vicente~\cite{fransizlarOE} based their method on the observation that
the movements of various patterns like smoke plumes produce
correlated temporal segments of gray-level pixels. They utilized
fractal indexing using a space-filling Z-curve concept along with
instantaneous and cumulative velocity histograms for possible smoke
regions. They made smoke decisions about the existence of smoke
according to the standard deviation, minimum average energy, and
shape and smoothness of these histograms. It is possible to include most of the currently available methods as sub-algorithms in the proposed framework and combine their decisions using the proposed EADF method.

Smoke at far distances ($>100$~m to the camera) exhibits different
spatio-temporal characteristics than nearby smoke and
fire~\cite{icip05}\nocite{icassp05}-\cite{patrec}. This demands
specific methods explicitly developed for smoke detection at far
distances rather than using nearby smoke detection methods described
in~\cite{eusipco2005}. The proposed approach is in accordance with the
`weak' Artificial Intelligence (AI) framework~\cite{Pavlidis}
introduced by Hubert~L.~Dreyfus as opposed to `generalized' AI.
According to this framework each specific problem in AI should be
addressed as an individual engineering problem with its own
characteristics~\cite{Dreyfus72},~\cite{Dreyfus92}.

\section{Building Blocks of Wildfire Detection Algorithm}
\label{sec:building_blocks}

Wildfire detection algorithm is
developed to recognize the existence of wildfire smoke within the
viewing range of the camera monitoring forestal areas. The proposed
wildfire smoke detection algorithm consists of five main
sub-algorithms: (i)~slow moving object detection in video,
(ii)~smoke-colored region detection, (iii)~wavelet transform based region smoothness detection, (iv)~shadow detection and elimination, (v)~covariance matrix based classification,
with decision functions, $D_1(x,n)$, $D_2(x,n)$, $D_3(x,n)$, $D_4(x,n)$ and
$D_5(x,n)$, respectively, for each pixel at location $x$ of every
incoming image frame at time step $n$. Computationally efficient
sub-algorithms are selected in order to realize a real-time wildfire
detection system working in a standard PC. The decision functions are combined in a linear
manner and the weights are determined according to the weight update mechanism described in Section~\ref{sec:weight_adaptation}.

Decision functions $D_i,~~ i=1,...,M$ of sub-algorithms do not
produce binary values $1$ (correct) or $-1$ (false), but they
produce real numbers centered around zero for each incoming sample $x$. If the
number is positive (negative), then the individual algorithm decides
that there is (not) smoke due to forest fire in the viewing range of
the camera. Output values of decision functions express the
confidence level of each sub-algorithm. Higher the value, the more
confident the algorithm.

First four sub-algorithms are described in detail in~\cite{toretez} which is available online at EURASIP webpage. We recently added the fifth sub-algorithm to our system. It is briefly reviewed below.

\subsection{Covariance Matrix Based Region Classification}
The fifth sub-algorithm deals with the classification of the smoke colored moving regions. A region covariance matrix~\cite{porikli} consisting of discriminative features is calculated for each region. For each pixel in the region, a 9-dimensional feature vector $z_k$ is calculated as follows:

 \begin{align} 
z_k=\Bigg[{x_1}\;{x_2}\;Y({x_1},{x_2})\;U({x_1},{x_2})\;V({x_1},{x_2}) \quad \quad  \quad\\ \nonumber
 \left| {\frac{{dY({x_1},{x_2})}}{{d{x_1}}}} \right| \;\left| {\frac{{dY({x_1},{x_2})}}{{d{x_2}}}} \right|\;\left| {\frac{{{d^2}Y({x_1},{x_2})}}{{dx_1^2}}} \right|\;\left| {\frac{{{d^2}Y({x_1},{x_2})}}{{dx_2^2}}} \right| \Bigg]^T \\ \nonumber
 \end{align} 

where $k$ is  the label of a  pixel, $(x_1 , x_2)$ is the  location of the pixel, $Y,U,V$ are the components of the representation of the pixel in YUV color space, $\frac{{dY(x_1,x_2)}}{{dx_1}}$ and $\frac{{dY(x_1,x_2)}}{{dx_2}}$ are the horizontal and vertical derivatives of the region respectively, calculated using the filter [-1 0 1], $\frac{{{d^2}Y(x_1,x_2)}}{{d{x_1^2}}}$ and $\frac{{{d^2}Y(x_1,x_2)}}{{d{x_2^2}}}$ are the horizontal and vertical second derivatives of the region calculated using the filter [-1 2 -1], respectively.

The feature vector for each pixel can be represented as follows: 
\begin{equation}
{z_k} = {[{z_k}(i)]^T}
\end{equation}
where, $z_k(i)$ is the $i_th$ entry of the feature vector. This feature vector is used to calculate the 9 by 9 covariance matrix of the regions using the fast covariance matrix computation formula~\cite{tuzel}:
\begin{equation}
{C_R} = [{c_R}(i,j)] = \left( {\frac{1}{{n - 1}}\left[ {\sum\limits_{k = 1}^n {{z_k}(i){z_k}(j)}  - Z_{kk} } \right]} \right)
\end{equation}
where
\begin{equation}\nonumber
Z_{kk}=\frac{1}{n}\sum\limits_{k = 1}^n {{z_k}(i)} \sum\limits_{k = 1}^n {{z_k}(j)}
\end{equation}
and $n$ is  the total number of pixels in the region and $c_R(i,j)$ is the $(i, j)$ the  component of the covariance matrix. 

The region covariance matrices are symmetric therefore we only need half of the elements of the matrix for classification. We also do not need the first 3 elements $c_R(1,1), c_R(2,1), c_R(2,2)$ when using the lower diagonal elements of the matrix, because these are the same for all regions. Then, we need a feature vector $f_R$ with $9\times10/2-3=42$ elements for each region. For a given region the final feature vector does not depend on the number of pixels in the region, it only depends on the number of features in $z_k$.

\begin{table}[!ht]
\caption{Confusion matrix of the training set}\label{table:svm}
\vspace{0.5cm}
\centering
 \begin{tabular}
{|c|l|c|c|}
\cline{3-4}
 \multicolumn{2}{c|}{} & \multicolumn{2}{|c|}{Predicted Labels} \\
\cline{3-4}
\cline{3-4}
 \multicolumn{2}{c|}{}&Not Smoke & Smoke\\
\hline
Actual&Not Smoke &   11342/(97.2)\% &     327/ (3.8\%)\\
\cline{2-4}
Labels&Smoke &   49/ (0.7\%) &  6962/(99.3\%)\\
\hline
\end{tabular}
\end{table}

\begin{figure}[!ht]
\centering

\subfigure[Negative training images.]{
   \includegraphics[width = 8.5cm] {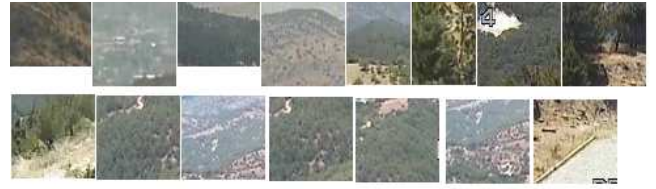}
 }

 \subfigure[Positive training images]{
   \includegraphics[width = 8.5cm] {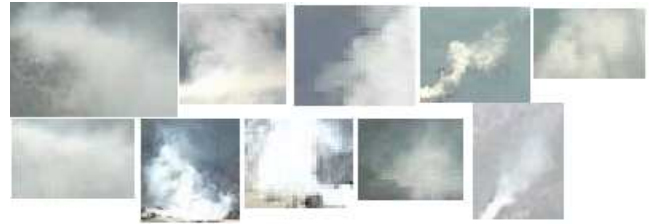} 
 } 
\caption{Positive and negative images from the training set.}
\label{myfigure}
\end{figure}

A Support Vector Machine (SVM) with RBF kernel is trained with the region covariance feature vectors of smoke regions in the training database. 18680 images are used to train the SVM. 7011 of the images are positive images which have actual smoke and the rest are negative images that do not have smoke. Sample positive and negative images are shown in Fig.~\ref{myfigure}. The confusion matrix for the training set is given in  Table~\ref{table:svm}. The success rate is $99.3\%$ for the positive images and $97.2\%$ for the negative images.

The LIBSVM~\cite{libsvm} software library is used to obtain the posterior class probabilities, $p_R=Pr(label=1|f_R)$, where $label=1$ corresponds to a smoke region. In this software library, posterior class probabilities are estimated by approximating the posteriors with a sigmoid function as in~\cite{platt}. If the posterior probability is larger than $0.5$ the label is 1 and the region contains smoke according to the covariance descriptor. The decision function for this sub-algorithm is defined as follows:
\begin{equation}\label{eq:svm}
D_5(x,n) = 2p_R-1
\end{equation}
where $0<p_R<1$ is the estimated posterior probability that the region contains smoke. In~\cite{porikli}, a distance measure based on eigenvalues are used to compare covariance matrices but we found that individual covariance values also provide satisfactory results in this problem.

As pointed above decision results of five sub-algorithms, $D_1$, $D_2$, $D_3$, $D_4$ and
$D_5$ are linearly combined to reach a final decision on a given
pixel whether it is a pixel of a smoke region or not. Morphological operations are applied to the detected pixels to mark the smoke regions. The number of connected smoke pixels should be larger than a threshold to issue an alarm for the region. If a false alarm is issued during training phase, the oracle gives feedback to the algorithm by declaring a no-smoke decision value ($y=-1$) for the false alarm region. Initially, equal weights
are assigned to each sub-algorithm.  There may be large variations between forestal areas and substantial temporal changes may occur within the same forestal region. As a result weights of individual sub-algorithms will evolve in a dynamic manner over time. 

In real-time operating mode the PTZ cameras are in continuous scan mode visiting predefined preset locations. In this mode constant monitoring from the oracle can be relaxed by adjusting the weights for each preset once and then using the same weights for successive classifications. Since the main issue is to reduce false alarms, the weights can be updated when there is no smoke in the viewing range of each preset and after that the system becomes autonomous. The cameras stop at each preset and run the detection algorithm for some time before moving to the next preset. By calculating separate weights for each preset we are able to reduce false alarms. 

\section{Experimental Results}
\label{sec:experimental} 
\subsection{Experiments on wildfire detection}
\begin{figure} [ht]
\centerline{\includegraphics*[width=8.5cm]{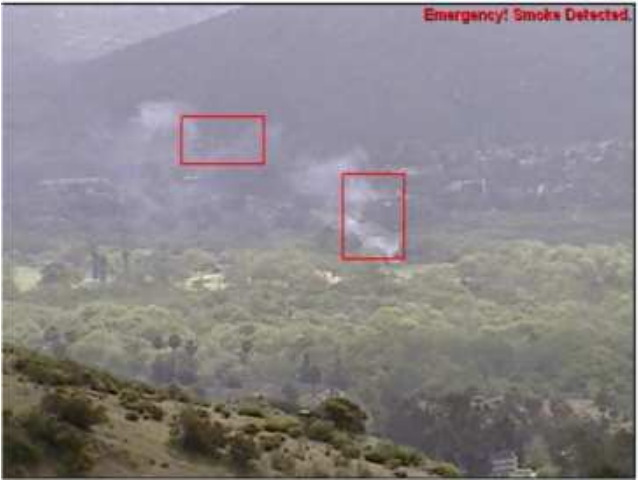}}
  \caption{A snapshot from an independent test of the system by the Regional Technology Clearing
House of San Diego State University in California in April 2009. The system successfully detected the
test fire and did not produce any false alarms. The detected smoke regions are marked with bounding rectangles.}\label{fig:san_diego}
\end{figure}

The proposed wildfire detection scheme with
entropy functional based active learning method is implemented on a PC with an
Intel Core Duo CPU 2.6GHz processor and tested with forest
surveillance recordings captured from cameras mounted on top of
forest watch towers near Antalya and Mugla provinces in Mediterranean region in Turkey. The weather is stable with sunny days throughout entire summer in Mediterranean. If it happens to rain there is no possibility of forest fire.  {\em
The installed system successfully detected three forest fires in the
summer of 2008}. The system is also independently tested by the Regional Technology Clearing
House of San Diego State University in California in April 2009 and it detected the
test fire and did not produce any false alarms. A snapshot from this test is presented in Fig.~\ref{fig:san_diego}. It also detected another forest fire in Cyprus in 2010.
The proposed EADF strategy is compared with the, projection onto convex sets (POCS) based algorithm and the
universal linear predictor (ULP) scheme proposed by
Singer and
Feder~\cite{singer_feder}. The ULP adaptive filtering method is modified to the wildfire detection problem in an online learning framework. In the ULP
scheme, decisions of individual algorithms are linearly combined
similar to Eq.~\ref{eq:y_hat_lms} as follows:
\begin{equation}
\hat{y}_u(x,n) = \sum_{i} v_i(n)D_i(x,n)
\end{equation}
where the weights, $v_i(n)$, are updated according to the ULP
algorithm, which assumes that the data (or decision values
$D_i(x,n)$, in our case) are governed by some unknown probabilistic
model $P$~\cite{singer_feder}. The objective of a universal
predictor is to minimize the expected cumulative loss. An explicit
description of the weights, $v_i(n)$, of the ULP algorithm is given
as follows:
\begin{equation}
v_i(n+1) = \frac{exp(-\frac{1}{2c}\ell(y(x,n),D_i(x,n)))}{\sum_j
exp(-\frac{1}{2c}\ell(y(x,n),D_j(x,n)))}
\end{equation}
where $c$ is a normalization constant and the loss function for the
$i$-th decision function is:
\begin{equation}
\ell(y(x,n),D_i(x,n)) = [y(x,n)-D_i(x,n)]^2
\end{equation}
The constant $c$ is taken as $4$ as indicated
in~\cite{singer_feder}. The universal predictor based algorithm is
summarized in Algorithm~\ref{fig:algo_universal}.

\begin{algorithm}[H]

\begin{algorithmic}
\STATE{Universal Predictor(x,n)}

\FOR{$i$ = 1 to M}
\STATE{$\ell(y(x,n),D_i(x,n)) = [y(x,n)-D_i(x,n)]^2$} 
\STATE{$v_i(n+1) =
\frac{exp(-\frac{1}{2c}\ell(y(x,n),D_i(x,n)))}{\sum_jexp(-\frac{1}{2c}\ell(y(x,n),D_j(x,n)))}$}
\ENDFOR
 \STATE{$\hat{y}_u(x,n) =\sum_{i} v_i(n)D_i(x,n)$} 
\IF{$\hat{y}_u(x,n) \geq 0$} 
\STATE{return 1} 
\ELSE
 \STATE{return -1}
 \ENDIF 
\end{algorithmic}
\caption{The pseudo-code for the universal predictor}
\label{fig:algo_universal}
\end{algorithm}

The POCS based scheme, the ULP based scheme, the EADF based scheme,
and the non-adaptive approach with fixed weights are compared in the following experiments. In
Tables~\ref{forest_comparison_table_detection}~and~\ref{forest_comparison_table_false_alarms},
6-hour-long forest surveillance recordings containing actual forest
fires and test fires as well as video sequences with no fires are
used.

\begin{figure}[ht]
\centerline{\includegraphics*[width=8.5 cm]{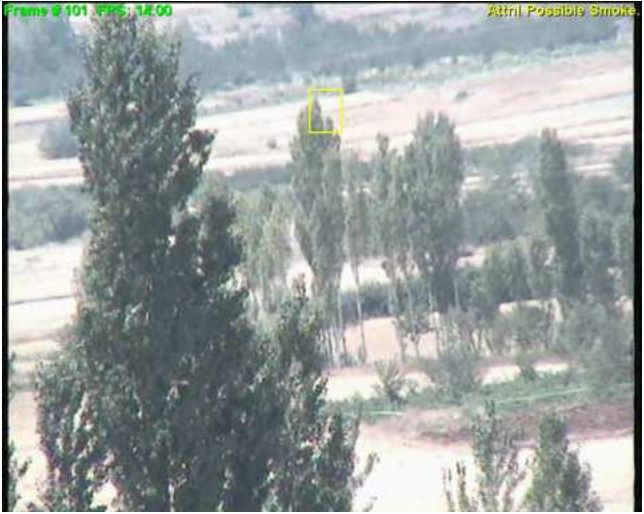}}
  \caption{False alarm from clip $V12$. Moving tree leaves in a forestal area cause a false alarm in an untrained algorithm with decision weights equal to $\frac{1}{5}$(depicted as a bounding box). The proposed algorithm does not produce a  false alarm in this video.}\label{fig:false_alarm}

\end{figure}

We have 7 test fire videos ranging
from 1~km to 4~km captured in Antalya and Mugla provinces in Mediterranean region in Turkey,
in the summers of 2007 and 2008. To the best of our knowledge this is the largest database of forest fire clips having the initial stages of wildfires. The database is also used by the European Commission funded project FIRESENSE~\cite{firesense}. All of the above mentioned decision
fusion methods detect forest fires within 20~seconds, as shown in
Table~\ref{forest_comparison_table_detection}. The detection rates
of the methods are comparable to each other. On the other hand, the
proposed adaptive fusion strategy significantly reduces the false
alarm rate of the system by integrating the feedback from the guard
(oracle) into the decision mechanism within the active learning
framework described in Section~\ref{sec:weight_adaptation}. 
In~Fig.~\ref{fig:false_alarm} a typical false alarm issued for
moving tree leaves (which cause the white background to appear as moving smoke), by an untrained algorithm with decision weights equal to $\frac{1}{5}$ is shown from the clip $V12$. The proposed
algorithm does not produce a false alarm in this video.

\begin{table}[ht]
\caption{Frame numbers at which an alarm is issued with different
methods for wildfire smoke captured at various ranges and fps. It is
assumed that the smoke starts at frame $0$.} \vspace*{0.0cm}
\begin{center}
{\small
\renewcommand\arraystretch{1.5}
 \renewcommand\tabcolsep{2pt} 
\begin{tabular}{|c|c|c|c|c|c|c|}
\hline       {Video}  &  {Range} & {Capture} & \multicolumn{4}{c|}{Frame number of first alarm} \\ \cline{4-7}{Sequence} & (km)  & Frame Rate& POCS  & Universal & Fixed  & EADF  \\ 
{}    &  & (fps) & Based &  & Weights & Based \\ \hline
{V1}  & 1 & 7  & 47  & 48  & 47    & 42   \\ \hline
{V2}  & 3 & 7 & 135 & 142 & 141   & 125   \\ \hline
{V3}  & 3 & 7 & 130 & 101 & 37    & 134  \\ \hline
{V4}  & 4 & 25 & 160 & 154 & 70    & 150   \\ \hline
{V5}  & 3 &9  & 65  & 55  & 57    & 56   \\ \hline
{V6}  & 2 & 5  & 70  & 74  & 76    & 75   \\ \hline
{V7}  & 2 & 5  & 93  & 74  & 41    & 83   \\ \hline
{Average} &- &-& 112.85 & 92.57 & 67  & 97.85   \\ \hline
\end{tabular}
}
\end{center}
\label{forest_comparison_table_detection}
\end{table}

\begin{table}[ht]
\caption{Average squared pixel errors issued by different methods to
video sequences without any wildfire smoke.} \vspace*{0.0cm}
\begin{center}
{\small
\renewcommand\arraystretch{1.5}
 \renewcommand\tabcolsep{2pt} 
\begin{tabular}{|c|c|c|c|c|c|c|}
\hline {} & {Frame} & {Video} & \multicolumn{4}{c|}{\multirow{2}{*}{Average Errors ($\times 10^{-3}$) }}\\
Video  & {Rate}  & {Duration}  & \multicolumn{4}{c|}{} \\
\cline{4-7}
{Sequence} & (fps) & (sec.)  & POCS & Universal & Fixed & EADF\\
&  &  & Based &{}  & Weights & Based \\ \hline
 {V8}  	 &7 &480 	       &  7.0076  & 94.6995  &138.2102 &   9.3712  \\ \hline 
  {V9}   	&25&300       &8.3375    & 38.2390 &  54.9168   & 5.5494   \\ \hline 
  {V10}  	&25 &600      &8.9892     & 77.5699 & 101.7512  &  3.2637   \\ \hline
  {V11}  	&10 &900      &5.2054     & 23.6602  & 30.0455   & 2.4314   \\ \hline 
  {V12}       &7 &60       &15.5350    & 98.0371 & 136.8163  & 12.1520   \\ \hline 
   {Average}     &-&- &   9.0149 &  66.4411  & 92.3480 &   6.5535     \\ \hline

\end{tabular}
}
\end{center}
\label{forest_comparison_table_false_alarms}
\end{table}

\begin{figure*}[!ht]
\centerline{\includegraphics*[width=6.5in]{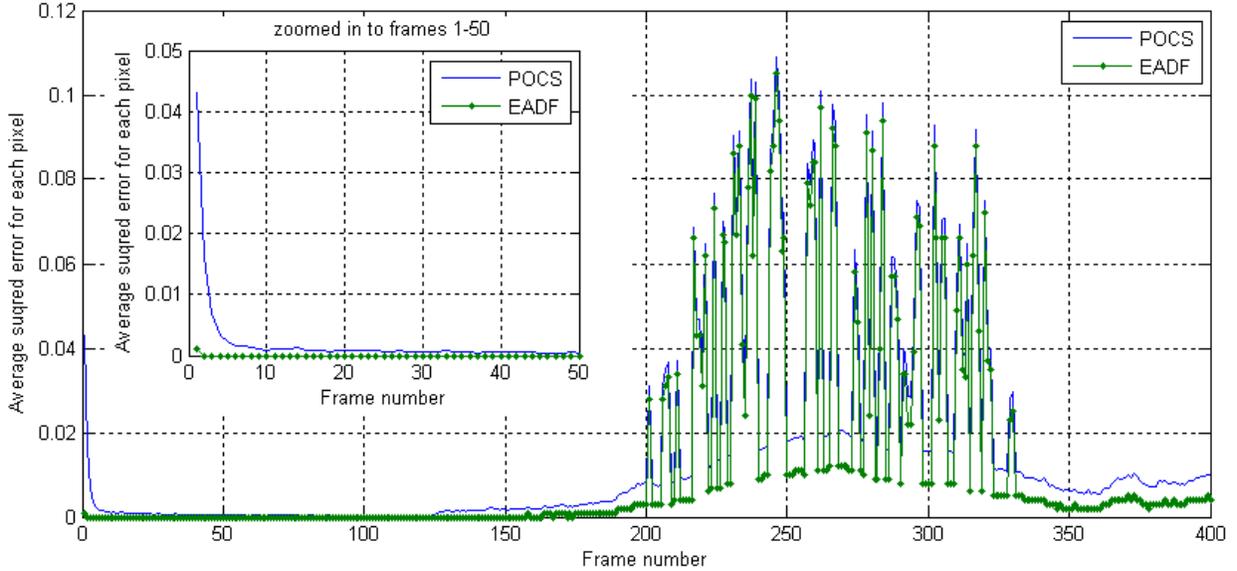}}
  \caption{Average squared pixel errors for POCS and EADF based algorithms for the video seuqence $V12$.}\label{fig:errors}
\end{figure*}

The proposed method produces the lowest average error in our data set. A set of video clips containing moving cloud shadows and other moving regions that usually cause false alarms is used to generate
Table~\ref{forest_comparison_table_false_alarms}. These video clips
are especially selected. The table shows the average pixel classification error for each method. The average pixel error for a video sequence $v$ is calculated as follows:
\begin{equation}
\bar{E}(v) = \frac{1}{F_I}\sum_{n=1}^{F_I} (\frac{e_n}{N_I})
\end{equation}
where $N_I$ is the total number of pixels in the image frame, $F_I$ is the number of frames in the video sequence, $e_n$ is the sum of the squared errors for each classified pixel in image frame $n$. Except for one video sequence EADF based method has the lowest pixel classification error.

In Fig.~\ref{fig:errors}, the squared pixels errors of POCS and EADF based schemes are compared for the video clip $V12$. The weights are updated until 125th frame for both algorithms. The POCS based algorithm has an initial stage until 30th frame where the error gradually drops to zero, whereas EADF algorithm converges after only 2 frames. The tracking performance of the EADF algorithm is also better than the POCS based algorithm which can be observed after the frame number 180 at which some of the sub-algorithms issue false alarms.

The software is currently being used in $59$ forest watch towers in Turkey. 

\subsection{Experiments on a UCI Dataset}
The proposed method is also tested with a dataset from UCI (University of California, Irvine) machine learning repository to evaluate the performance of the algorithm in combining different classifiers. In the wildfire detection case the image data arrives sequentially and the decision weights are updated in real-time. On the other hand the UCI data sets are fixed. Therefore the dataset is divided into two parts. The first part is used for training.

During the training phase, weights of different classifiers are determined using the EADF update method. In testing stage the fixed weights obtained from the training stage are used to combine the classifier decisions which process the data in a sequential manner because both the POCS and EADF frameworks assume that the new data arrive in a sequential manner.

The test is performed on the ionosphere data from UCI machine learning repository that consists of radar measurements to detect the existence of free electrons that form a structure in the atmosphere. The electrons that show some kind of structure in the ionosphere return ``Good'' responses, the others return ``Bad'' responses. There are 351 samples with 34-element feature vectors that are obtained by passing the radar signals through an autocorrelation function. In~\cite{ion}, the first 200 samples are used as training set to classify the remaining 151 test samples. They obtained \% 90.7 accuracy with a linear perceptron, \% 92 accuracy with a non-linear perceptron, and \% 96 accuracy with a back propagation neural network.

For this test SVM, k-nn (k-Nearest Neighbor) and NCC (normalized cross-correlation) classifiers are used. Also, in this classification the decision functions of these  classifiers produce binary values with 1 corresponding to ``Good'' classification and -1 corresponding to ``Bad'' classification rather than scaled posterior probabilities in the range $[-1,1]$. 

The accuracies of the sub-algorithms and EADF are shown in Table~\ref{iones}. The success rates of the proposed EADF and POCS methods are both \% 98.01 which is higher than all the sub-algorithms. Both the entropic projection and orthogonal projection based algorithms converge to a solution in the intersection of the convex sets. It turns out that they both converge to the same solution in this particular case. This is possible when the intersection set of convex sets is small. The proposed EADF method is actually developed for real-time application in which data arrives sequentially. This example is included to show that the EADF scheme can be also used in other datasets. It may be possible to get better classification results with other classifiers in this fixed UCI dataset.

\begin{table}[ht]
\caption{Accuracies of sub-algorithms and EADF on ionosphere dataset.} 
\begin{center}
{\small
\renewcommand\arraystretch{1.5}
 \renewcommand\tabcolsep{2pt} 
\begin{tabular}{|c|c|c|c|c|c|}
\hline {\multirow{2}{*}{Data}}  & \multicolumn{5}{c|}{Success Rates (\%)} \\
\cline{2-6}
        &  SVM     & k-nn (k=4) & NCC       & POCS  & EADF  \\  \hline
{Train} &  100.0   & 91.50      & 100.0     & 100.0 & 100.0 \\  \hline
{Test}  &  94.03   & 97.35      & 91.39     & 98.01 & 98.01 \\  \hline
\end{tabular}
}
\end{center}
\label{iones}
\end{table}
\section{Conclusion}
\label{sec:conclusion} 
An entropy functional based online adaptive decision fusion (EADF) is proposed for image analysis and 
computer vision applications with drifting concepts. In this framework,
it is assumed that the main algorithm for a specific application is composed of several sub-algorithms each of which yielding its own decision as a real number centered around zero representing its confidence level. Decision values
are linearly combined with weights which are updated online by performing non-orthogonal e-projections onto convex sets describing sub-algorithms. This general framework is applied to a real computer vision problem of wildfire detection. The proposed adaptive decision fusion
strategy takes into account the feedback from guards of forest watch
towers. Experimental results show that the learning duration is
decreased with the proposed online adaptive fusion scheme. It is
also observed that error rate of the proposed method
is the lowest in our data set, compared to universal linear
predictor (ULP) and the projection onto convex sets (POCS) based schemes.

The proposed framework for decision fusion is
suitable for problems with concept drift. At each stage of the
algorithm, the method tracks the changes in the nature of the
problem by performing an non-orthogonal e-projection onto a hyperplane
describing the decision of the oracle.



\section*{Acknowledgment}
This work was supported in part by the Scientific and Technical
Research Council of Turkey, TUBITAK, with grant no. 106G126 and
105E191, in part by European Commission 6th Framework Program
with grant number FP6-507752 (MUSCLE Network of Excellence Project) and in part by FIRESENSE (Fire Detection and Management through a Multi-Sensor Network for the Protection of Cultural Heritage Areas from the Risk of Fire and Extreme Weather Conditions, FP7-ENV-2009-1–244088-FIRESENSE) .

\bibliographystyle{IEEEbib}
\footnotesize
\bibliography{refer}
\end{document}